\documentclass[conference]{IEEEtran}
\IEEEoverridecommandlockouts

\usepackage{cite}
\usepackage{amsmath,amssymb,amsfonts}
\usepackage{algorithmic}
\usepackage{graphicx}
\usepackage{textcomp}
\usepackage{xcolor}
\usepackage[hidelinks]{hyperref}
\usepackage{academicons}
\usepackage{orcidlink}
\usepackage{balance}
\usepackage{booktabs}
\usepackage{tabularx}
\usepackage{array}
\usepackage{tikz}
\usetikzlibrary{trees}

\usepackage{enumitem}
\setlist{nosep}

\usepackage{authblk}

\usepackage[T5]{fontenc}
\usepackage{balance}
\usepackage[utf8]{inputenc}
\DeclareTextSymbolDefault{\DH}{T1}
\usepackage{lmodern} 
\makeatletter
\newcommand{\printfnsymbol}[1]{%
  \textsuperscript{\@fnsymbol{#1}}%
}
\makeatother

\newcommand{\comment}[1]{\textcolor{black}{#1}}

\def\BibTeX{{\rm B\ker†n-.05em{\sc i\kern-.025em b}\kern-.08em
    T\kern-.1667em\lower.7ex\hbox{E}\kern-.125emX}}
\begin{document}

\title{Toward Content-based Indexing and Retrieval of Head and Neck CT with Abscess Segmentation}


\author[1, 2, 3]{Thao Thi Phuong Dao\orcidlink{0000-0002-0109-1114}}

\author[1, 2, 4]{Tan-Cong Nguyen\orcidlink{0000-0001-8834-8092}}

\author[1, 2]{Trong-Le Do\orcidlink{0000-0002-2906-0360}}

\author[3]{Truong Hoang Viet\orcidlink{0009-0007-9433-0017}}

\author[3]{\authorcr Nguyen Chi Thanh\orcidlink{0009-0002-8108-1502}}

\author[3]{Huynh Nguyen Thuan\orcidlink{0009-0002-6437-918X}}

\author[3]{Do Vo Cong Nguyen\orcidlink{0009-0005-9505-9404}}
\author[5]{Minh-Khoi Pham\orcidlink{0000-0003-3211-9076}}
\author[1, 2]{\authorcr Mai-Khiem Tran\orcidlink{0000-0001-5460-0229}}
\author[1, 2]{Viet-Tham Huynh\orcidlink{0000-0002-8537-1331}}
\author[1, 2]{Trong-Thuan Nguyen\orcidlink{0000-0001-7729-2927}}
\author[1, 2]{Trung-Nghia Le\orcidlink{0000-0002-7363-2610}}
\author[3]{\authorcr  Vo Thanh Toan\orcidlink{0000-0003-3511-9987}}
\author[7]{Tam V. Nguyen\orcidlink{0000-0003-0236-7992}}
\author[1, 2, 6]{Minh-Triet Tran\orcidlink{0000-0003-3046-3041}$^\dagger$\thanks{$\dagger$Corresponding authors and advisors.}}
\author[8, 2, 3]{Thanh Dinh Le\orcidlink{0009-0009-3153-085X}$^\dagger$}

\affil[1]{University of Science, VNU-HCM, Ho Chi Minh City, Vietnam}
\affil[2]{Vietnam National University, Ho Chi Minh City, Vietnam}
\affil[3]{Thong Nhat Hospital, Ho Chi Minh City, Vietnam}
\affil[4]{University of Social Sciences and Humanities, VNU-HCM, Ho Chi Minh City, Vietnam}
\affil[5]{Dublin City University, Dublin, Ireland}
\affil[6]{John von Neumann Institute, Ho Chi Minh City, Vietnam}
\affil[7]{University of Dayton, Dayton, Ohio, United States}
\affil[8]{University of Health Sciences, VNU-HCM, Ho Chi Minh City, Vietnam}

\maketitle


\begin{abstract}
Abscesses in the head and neck represent an acute infectious process that can potentially lead to sepsis or mortality if not diagnosed and managed promptly. Accurate detection and delineation of these lesions on imaging are essential for diagnosis, treatment planning, and surgical intervention. 
In this study, we introduce \textbf{AbscessHeNe}, a curated and comprehensively annotated dataset comprising 4,926 contrast-enhanced CT slices with clinically confirmed head and neck abscesses. The dataset is designed to facilitate the development of robust semantic segmentation models that can accurately delineate abscess boundaries and evaluate deep neck space involvement, thereby supporting informed clinical decision-making.
To establish performance baselines, we evaluate several state-of-the-art segmentation architectures, including CNN, Transformer, and Mamba-based models. The highest-performing model achieved a Dice Similarity Coefficient of 0.39, Intersection-over-Union of 0.27, and Normalized Surface Distance of 0.67, indicating the challenges of this task and the need for further research.
\comment{
Beyond segmentation, AbscessHeNe is structured for future applications in \textbf{content-based multimedia indexing and case-based retrieval}. Each CT scan is linked with pixel-level annotations and clinical metadata, providing a foundation for building intelligent retrieval systems and supporting knowledge-driven clinical workflows. The dataset will be made publicly available at  \url{https://github.com/drthaodao3101/AbscessHeNe.git}.}

\end{abstract}

\begin{IEEEkeywords}
Abscess Dataset, Head \& Neck CT scan, Deep Learning, Medical Image Segmentation
\end{IEEEkeywords}

\section{Introduction}
\label{sec:int}
Head and neck abscesses are a critical subset of deep neck infections, characterized by localized collections of pus within the deep fascial compartments of the neck \cite{yang2008analysis}. These infections often arise as a complication of cellulitis following infections of the oral cavity, pharynx, or upper respiratory tract. They may also be precipitated by trauma, congenital anomalies, or the ingestion of foreign bodies \cite{dou2024risk}. Although the overall incidence of deep neck infections has declined in recent decades due to advancements in antibiotic therapy, diagnostic imaging, and intensive care management, abscess formation within the head and neck region remains a significant clinical concern. A quick and accurate diagnosis is essential, as delayed intervention may allow for rapid disease progression within the complex anatomical spaces of the neck. It can result in life-threatening complications such as airway obstruction, mediastinal spread, septic shock, or disseminated intravascular coagulation (DIC) \cite{vieira2008deep}.
However, diagnosing and treating head and neck abscesses remains highly challenging due to the intricate and variable anatomy of the deep cervical spaces. The presence of vital structures such as major blood vessels, cranial nerves, and the airway, combined with the distortion of normal anatomy by inflammation and edema, complicates imaging interpretation and surgical planning \cite{pandey2019perspective,lim2021dangerous}. Therefore, accurate delineation of abscess boundaries is critical to minimize surgical risks, reduce operative morbidity, and optimize patient outcomes.

Contrast-enhanced computed tomography (CECT) is widely regarded as the imaging modality for evaluating head and neck abscesses. CECT offers high-resolution visualization of both soft tissue and osseous structures, providing essential information on lesions, vascular involvement, and potential airway obstruction \cite{yonetsu1998deep,hagelberg2022diagnostic}. Nevertheless, interpreting CECT images requires substantial expertise, and even experienced clinicians may encounter difficulties in precisely identifying abscess margins, particularly in complex or multiloculated cases.

In recent years, artificial intelligence (AI), particularly deep learning (DL), has demonstrated remarkable progress in medical image analysis \cite{phung2022disease, ho2023multiclass, rayed2024deep, huynh2024dermai,nguyen2023manet}. Models have shown promise in automating lesion detection, enhancing diagnostic accuracy, and supporting informed decision-making across various healthcare domains. Within otolaryngology and head and neck surgery, AI-based algorithms have been explored for applications such as tumor segmentation, lymph node evaluation, and identification of anatomical landmarks \cite{tama2020recent,wang2021deep,luo2024multicenter, le2024medgraph}. However, the application of AI for segmentation of head and neck abscesses remains underexplored, primarily due to the lack of dedicated annotated datasets for algorithm development and validation.

To address this gap, we present a novel dataset, AbscessHeNe, of head and neck abscesses derived from CECT scans. Specifically, this dataset comprises 4,926 image–mask pairs derived from 47 patients for semantic segmentation, accompanied by corresponding report files that provide detailed descriptions of each patient’s CECT scans. 
The results of the best-performing model indicate that segmentation on our dataset remains challenging and warrants further research to enhance model performance. Our objective is to enable precise identification of abscess boundaries, thereby facilitating rapid diagnosis, guiding surgical planning, and reducing complications. 

This work also aims to move beyond segmentation toward content-based indexing and retrieval (CBIR) of head and neck CT. In our work, high-quality lesion masks serve as semantic anchors for representation learning. Features extracted from the segmented abscess regions, augmented by structured metadata (e.g., location, morphology descriptors, drainage, and outcome notes), enable indexing of cases by their content rather than by study-level tags alone. This structure supports several retrieval modes that are clinically meaningful: (i) query-by-image to find visually similar abscesses, (ii) query-by-mask to emphasize boundary shape and extent, and (iii) mixed queries that combine image features with key metadata (e.g., ``multiloculated parapharyngeal abscess with airway deviation''). Although we do not claim retrieval results in this study, AbscessHeNe is organized to facilitate future CBIR research, including evaluation with metrics such as Recall@K and mean Average Precision, and integration into decision support workflows where similar prior cases can inform drainage planning and risk assessment.


Our contributions are as follows:
\begin{itemize}
    \item We introduce \textbf{AbscessHeNe}, the first publicly available annotated dataset dedicated to head and neck abscess segmentation on CECT scans, addressing a critical gap in the analysis of non-oncologic lesions.
    \item We conduct a comprehensive benchmark of multiple state-of-the-art semantic segmentation models, spanning CNN, Transformer, and Mamba-based architectures, to establish strong performance baselines.
    \item We provide a detailed analysis of segmentation performance, with a focus on clinically relevant factors such as boundary precision and deep neck space involvement, highlighting its real-world applicability.
    \item \comment{We design AbscessHeNe with extensibility for \textbf{content-based indexing and case-based retrieval}, where each CT scan is linked to structured lesion annotations and clinical metadata, laying the groundwork for multimedia retrieval and clinical decision support applications.}
\end{itemize}

\section{Related works}
\label{sec:rel}

\begin{figure*}[t]
  \centering
    \includegraphics[width=0.9\textwidth,trim={0cm 0cm 0cm 0cm}, clip]{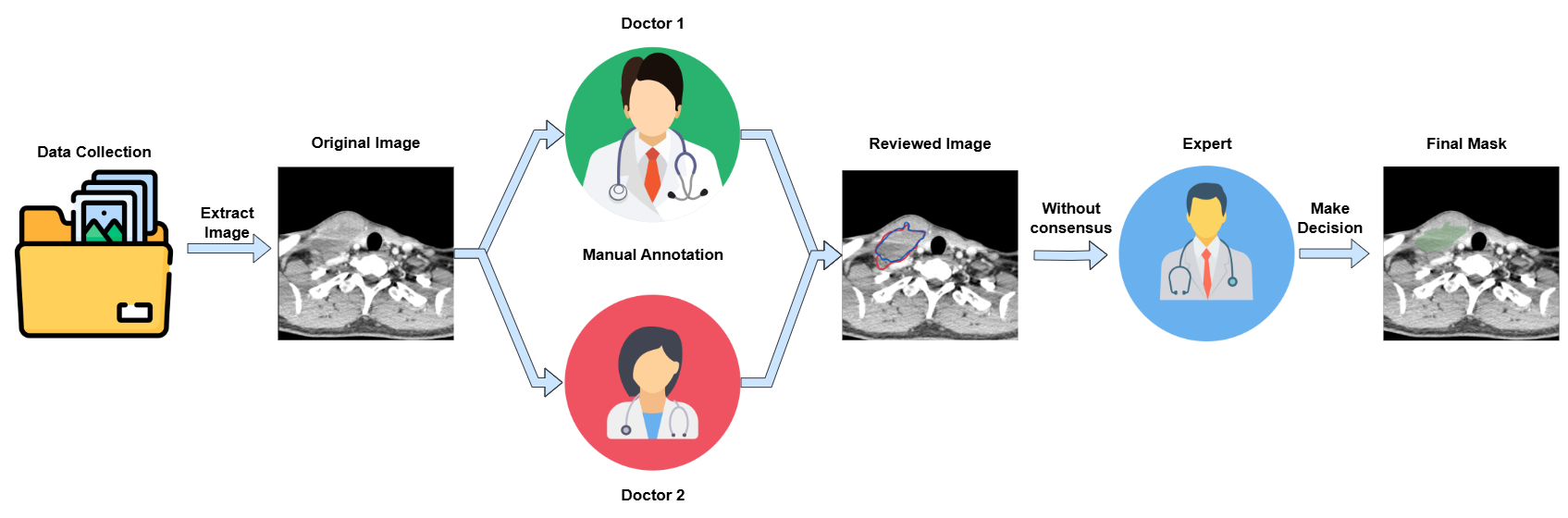}
    \vspace{-3mm}
    \caption{Comprehensive pipeline for building AbscessHeNe dataset.}
    \vspace{-5mm}
    \label{fig:process}
\end{figure*}

\subsection{Medical Image Segmentation and Retrieval}
In the field of medical image semantic segmentation, DL models have rapidly evolved into three main paradigms, including convolutional neural networks, Transformer, and, more recently, Mamba-based models. The following provides an overview of representative models.

Convolutional Neural Networks (CNNs) have long been the backbone of semantic segmentation in medical imaging due to their ability to capture local spatial details. The U-Net \cite{ronneberger2015u}  architecture remains one of the most influential designs, employing an encoder–decoder structure with skip connections to preserve fine-grained information during upsampling. Various solutions have also been proposed in this direction, such as U-Net++ \cite{zhou2018unet++}, ResUNet \cite{diakogiannis2020resunet}, or ResUNet++ \cite{jha2019resunet++}.

Although CNNs excel at modeling local patterns, their limited receptive field hampers global context modeling. Transformer-based architectures, leveraging self-attention, have addressed this limitation by capturing long-range dependencies more effectively. TransUNet \cite{chen2021transunet} pioneered the hybrid approach by combining CNN encoders with Transformer modules to jointly learn local and global semantics while maintaining a U-shaped decoding path. UNETR \cite{hatamizadeh2022unetr} fully replaced the encoder with a multi-layer Transformer, treating the input as a sequence of patches for holistic representation learning. However, such pure Transformer models incur significant computational cost, particularly for high-resolution 3D data. To mitigate this, Swin-UNETR \cite{hatamizadeh2021swin} employed the Swin Transformer \cite{liu2021swin} with shifted windows, enabling scalable self-attention with linear complexity relative to image size.

Recently, state space models such as Mamba \cite{gu2023mamba} have emerged as a promising alternative to Transformers, offering comparable long-range modeling capabilities with linear complexity. U-Mamba \cite{ma2024u} introduced this paradigm to medical segmentation by integrating Mamba blocks into a U-Net framework, combining the strengths of CNNs for local feature extraction and Mamba’s efficient sequence modeling. VMamba \cite{liu2024vmamba} extended this idea by adopting a two-dimensional selective scan (Visual State Space - VSS) to capture spatial dependencies inherent in images better. Swin-UMamba \cite{liu2024swin} further demonstrated the benefits of combining large-scale pretraining with Mamba-based encoders and decoders, achieving strong performance with reduced parameters and FLOPs. Finally, VM-UNet \cite{ruan2024vm}, a fully state space model with VSS-based encoder and decoder, highlighted the potential of pure Mamba-based designs to deliver lightweight yet competitive segmentation performance.

For segmentation-related retrieval settings, Ogura et al. \cite{ogura2024similar} employ contrastive learning to retrieve masks from the source domain as shape templates, leveraging cross-domain anatomical consistency to improve single-domain generalization over a baseline. Gul et al. \cite{gul2025hybrid} propose a hybrid pipeline for multi-panel figures that first classifies figure type, then segments both regular and irregular layouts using projection profiles and morphological operations, enabling precise and fast sub-image indexing and retrieval. Complementing these efforts, Zhao et al. \cite{zhao2025retrieval} introduce a retrieval-augmented few-shot segmenter that queries DINOv2 features to find similar annotated samples, stores them in a memory bank, and conditions SAM2 via memory attention. This method achieves strong cross-modality performance without retraining or fine-tuning.

\subsection{Head and Neck CT Scans Datasets for Segmentation and Retrieval}
\newcolumntype{L}{>{\raggedright\arraybackslash}X}












In the domain of head and neck CT imaging, the availability of high-quality, publicly accessible datasets has played a pivotal role in advancing the development and benchmarking of DL models. To date, the majority of publicly released head and neck CT datasets have been designed with a focus on oncologic applications, particularly for the diagnosis and radiotherapy planning of head and neck cancer. While these datasets have catalyzed progress in automated segmentation of tumors and organs-at-risk (OARs), they fall short in covering the broader spectrum of space-occupying lesions, such as abscesses, that are clinically critical in the head and neck region.

The Cancer Imaging Archive (TCIA) \cite{clark2013cancer} offers a rich collection of multimodal cancer imaging data, including CT, MRI, and PET, along with associated clinical annotations. Several derivative datasets have been curated from TCIA resources, such as the Public Domain Database for Computational Anatomy (PDDCA) \cite{raudaschl2017evaluation} (48 CT images with 9 OARs), the TCIA Test \& Validation Radiotherapy CT Planning Scan dataset \cite{nikolov2021clinically} (31 CT images with 21 OARs), and the UaNet dataset \cite{tang2019clinically} (140 CT scans with 28 OARs).
Furthermore, other notable datasets include StructSeg \cite{li2020automatic}, comprising 50 CT scans annotated with 22 OARs, and RT-MRI \cite{head2019prospective}, containing 15 paired CT and MRI scans annotated with 23 OARs. More recently, the HaN-Seg dataset \cite{podobnik2023han} provided 56 paired CT and MRI scans with segmentations of 30 OARs. RADCURE \cite{welch2024radcure} introduced an unprecedentedly large-scale cohort of 3,346 patients, offering simulation CT images with annotations of primary tumors, metastatic lymph nodes, and 19 OARs, integrated with comprehensive clinical and demographic metadata. Similarly, the SegRap2023 Challenge dataset \cite{luo2025segrap2023} provided 200 pairs of contrast and non-contrast CT scans annotated for 45 OARs and 2 gross tumor volumes (GTVs) in nasopharyngeal carcinoma patients.


Rather than releasing datasets solely for standalone segmentation, recent studies pair segmentation masks with metadata to support retrieval for clinical decision support and case-based reasoning. Community benchmarks continue to catalyze progress in retrieval, such as the ImageCLEF medical retrieval challenges \cite{ionescu2025imageclef} and MedMNIST \cite{yang2023medmnist}. Nevertheless, head-and-neck resources, especially for abscess cases, that combine masks and metadata for retrieval remain limited.


\section{AbscessHeNe Dataset}
\label{sec:data}

\subsection{Clinical Definition}
\label{sec:define}

An abscess is an abnormal pathological area characterized by a localized collection of pus. Pus consists of a mixture of dead white blood cells, bacteria, necrotic tissue, and fluid. Therefore, an abscess lesion typically includes three components: (1) a central necrotic core composed of inflammatory and necrotic cells; (2) a peripheral zone rich in white blood cells; and (3) an encapsulating fibrous wall with vascular dilation and fibroblast proliferation.

\begin{figure}[t]
  \centering
    \includegraphics[width=0.5\textwidth,trim={0cm 0cm 0cm 0cm}, clip]{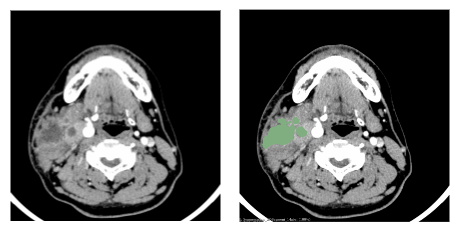}
    \vspace{-10mm}
    \caption{A sample of AbscessHeNe. Left-hand: original image and Right-hand: Image with mask (green)}
    \vspace{-5mm}
    \label{fig:sample}
\end{figure}

The radiologic appearance of an abscess may vary depending on its anatomical location, degree of infiltration, and local mass effect. On CECT imaging, abscesses often present as irregularly shaped lesions with variable sizes and directional spread. Internally, they are usually heterogeneous in density, possibly containing septations or small particulate components. The abscess wall is typically well-defined, varying in thickness, and demonstrates low attenuation due to the fluid and pus content, ranging from 0 to 45 Hounsfield units (HU). An example is shown in the left image of Figure \ref{fig:sample}.

\subsection{Dataset Construction}

\subsubsection{Data Collection}
CECT images used in this study were retrospectively collected from Thong Nhat Hospital, Vietnam, spanning from January 2019 to July 2024. All imaging data were acquired through the hospital’s Picture Archiving and Communication System (PACS), under the supervision of board-certified radiologists and otolaryngologists. An overview of the data acquisition and labeling process is illustrated in Figure \ref{fig:process}.
In this study, patients diagnosed with head and neck lesions confirmed as abscesses based on histopathological examination or positive culture results, followed by effective clinical treatment, were included. CT scans were obtained either at the time of initial diagnosis or during follow-up, in accordance with routine clinical protocols. All data were exported in the DICOM format.

CT scans were acquired using a Philips Brilliance iCT SP 128 CT Scanner (Philips Healthcare, Best, Netherlands), employing institutional protocols optimized for soft-tissue imaging. Non-ionic iodinated contrast agents were administered intravenously for patients with adequate renal function, enhancing the delineation of soft-tissue structures and abscess boundaries. Scanning parameters included a tube voltage of 120 kVp, tube current modulation between 200 and 300 mAs, slice thickness ranging from 1.5 to 3.0 mm, and an in-plane resolution of 0.5 × 0.5 mm, with a matrix size of 512. All scans were obtained in the supine position, covering from the skull base to the upper mediastinum to encompass the entire deep cervical spaces.
To ensure data quality, scans with severe motion artifacts, metallic interference, or incomplete anatomical coverage were excluded. The final dataset consisted of 4,926 CT slices from 47 cases of head and neck abscesses.

All imaging data were anonymized before analysis to protect patient privacy, in accordance with the ethical guidelines of Thong Nhat Hospital and the Declaration of Helsinki. Personally identifiable information, including patient identifiers, birth dates, and medical record numbers, was permanently removed. The study protocol was approved by the Institutional Ethics Committee of Thong Nhat Hospital (Approval No. 26/2023/BVTN-H\DJ{}Y\DJ{}).

\subsubsection{Data Annotation}
To ensure accurate and consistent delineation of abscess boundaries, a multi-stage annotation protocol was implemented involving experienced otolaryngologists and radiologists specializing in head and neck imaging. The annotation process was conducted in three sequential phases.

In the first phase, the dataset was distributed to two independent annotators. Each doctor manually annotated the abscess boundaries at the pixel level on each CT slice using specialized medical imaging software. For every case, the annotators carefully segmented the entire abscess region, including both the necrotic central cavity and any peripheral enhancing rims suggestive of inflamed tissue. 
In the second phase, the two annotators jointly reviewed cases with inconsistent annotations. Consensus discussions were conducted to resolve ambiguities and establish a unified segmentation. This collaborative review process helped minimize inter-annotator variability and improve label reliability. Unresolved cases were forwarded to the third phase.
In the third phase, an independent senior expert reviewed all segmentations and reports to ensure anatomical accuracy and inter-case consistency. The expert also adjudicated the final segmentations in disputed cases from Phase 2, providing the final ground-truth labels.

The AbscessHeNe dataset offers high-quality, pixel-wise abscess segmentations from 47 patients, comprising a total of 4,926 annotated CT slices in PNG format. The data were split into training and test sets in an 80:20 ratio, with 3,935 and 991 images, respectively. All images were resized to 512 × 512 pixels for spatial consistency. Figure \ref{fig:sample} shows a sample slice with ground truth annotation, where the abscess boundary is delineated.

\begin{table}[!t]
\centering
\caption{Comparative performance of state-of-the-art models on the AbscessHeNe dataset }
\label{tab:results}
\resizebox{\linewidth}{!}{
\begin{tabular}{l|cc|cccc}
\toprule
\textbf{Model} & \textbf{Param$\downarrow$} & \textbf{FLOPs$\downarrow$} & \textbf{DSC$\uparrow$} & \textbf{mIoU$\uparrow$} & \textbf{HD95$\downarrow$} & \textbf{NSD$\uparrow$} \\ \midrule
U-Net \cite{ronneberger2015u}           & 31.04 M          & 54.74 G         & 0.35          & 0.25          & 28.58          & 0.57           \\
TransUNet \cite{chen2021transunet}      & 105.32 M         & 32.23 G         & 0.27          & 0.19          & 30.52          & 0.59           \\
UNETR \cite{hatamizadeh2022unetr}       & 115.94 M         & 25.70 G         & 0.22          & 0.14          & 28.14          & 0.59           \\
Swin UNETR \cite{hatamizadeh2021swin}   & \textbf{25.14 M} & 19.15 G         & 0.22          & 0.14          & 31.23          & 0.62           \\
Swin-UMamba \cite{liu2024swin}          & 59.89 M          & 43.95 G         & 0.26          & 0.17          & 27.05          & 0.66           \\
VM-UNet \cite{ruan2024vm}               & 27.43 M          & \textbf{4.11 G} & \textbf{0.39} & \textbf{0.27} & \textbf{21.27} & \textbf{0.67}  \\ \bottomrule
\end{tabular}
}
\vspace{-5mm}
\end{table}

\section{Benchmark Suite}

\subsection{Segmentation Baselines}

We evaluated a range of architectures encompassing various deep learning paradigms for medical image segmentation. The models for evaluation include U-Net \cite{ronneberger2015u}, TransUNet \cite{chen2021transunet}, UNETR \cite{hatamizadeh2022unetr}, Swin-UNETR \cite{hatamizadeh2021swin}, Swin U-Mamba \cite{liu2024swin}, and VM-UNet \cite{ruan2024vm}. These architectures provide a comprehensive benchmark to assess segmentation performance on our dataset.

We implement baselines using the representative PyTorch platform on a PC equipped with an NVIDIA RTX A5000 GPU 24 GB. We employ corresponding optimizer techniques, utilizing an initial learning rate and a decay strategy, which are the default settings for training each model. These models are trained for 300 epochs on our dataset with a batch size of 16. In the training process, all images were resized to 256×256 pixels and augmented through random vertical or horizontal flipping and rotation to enhance data diversity.

Computational efficiency was assessed by the number of parameters (in millions) and floating-point operations per second (FLOPs, in gigaflops). Segmentation performance on the AbscessHeNe dataset was measured using the Dice Similarity Coefficient (DSC), mean Intersection over Union (mIoU), 95th percentile Hausdorff Distance (HD95), and Normalized Surface Distance (NSD) \cite{reinke2021common}.

%

\subsection{Quantitative  Comparison}

Table \ref{tab:results} summarizes the quantitative performance of the evaluated models on the AbscessHeNe dataset. Among the CNN-based and Transformer-based baselines, U-Net achieved a DSC of 0.35, outperforming TransUNet, UNETR, and Swin-UNETR in segmentation accuracy. Notably, the Mamba-based architectures demonstrated superior performance, with VM-UNet attaining the best overall results: a DSC of 0.39, mIoU of 0.27, HD95 of 21.27, and NSD of 0.67, while maintaining a low computational cost with 4.11 G FLOPs. This indicates that VM-UNet effectively balances segmentation accuracy and efficiency, surpassing prior state-of-the-art methods on this task.

\subsection{Qualitative Comparison}

\begin{figure*}[t]
  \centering
    \includegraphics[width=0.8\textwidth,trim={0cm 6cm 0cm 0cm}, clip]{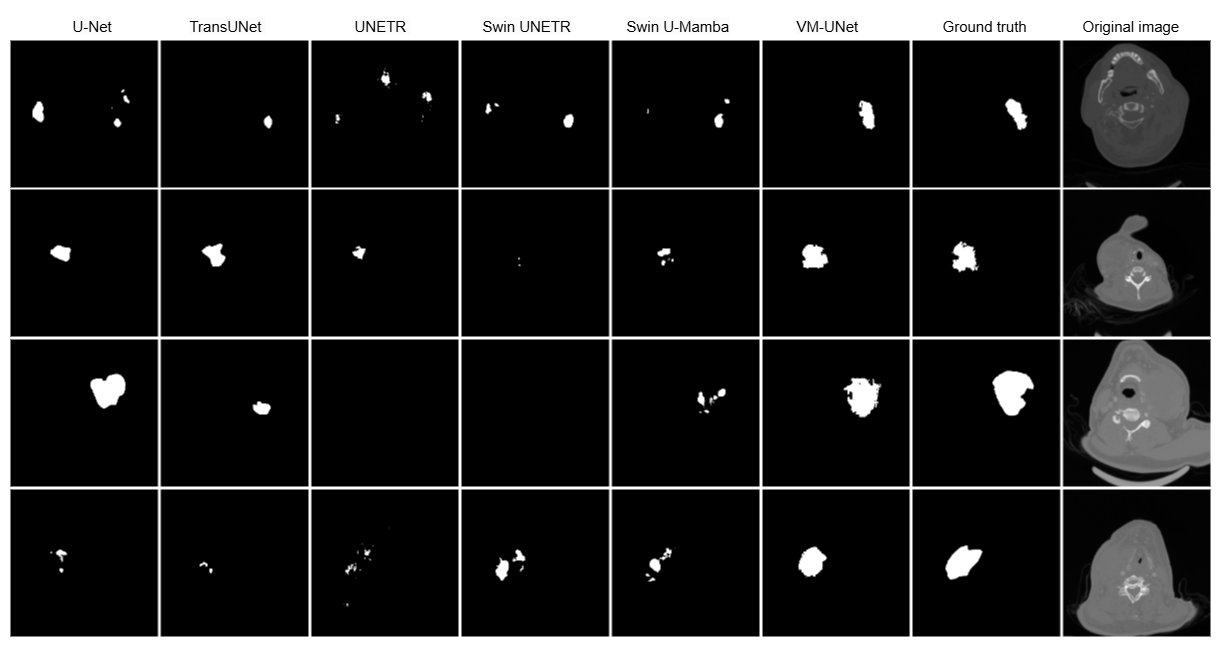}
    \caption{Visualization of results from the different models on our AbscessHeNe dataset for segmentation.}
    \label{fig:visualize}
    
\end{figure*}

The qualitative performance of the evaluated models is illustrated in Figure~\ref{fig:visualize}. As discussed in Section~\ref{sec:define}, the intrinsic characteristics of abscess lesions, namely, indistinct boundaries and heterogeneous internal density, pose significant challenges for accurate delineation. In the first row of Figure~\ref{fig:visualize}, most models exhibit false positive predictions, except for VM-UNet. In the second and third rows, the VM-UNet, U-Net, and TransUNet produce segmentation results more closely aligned with the ground truth, consistent with the quantitative findings presented in Table~\ref{tab:results}. Finally, the last row highlights the overall difficulty of the AbscessHeNe dataset, underscoring the need for future research to develop more robust algorithms for this segmentation task.

\subsection{Retrieval Prototype}

\begin{figure*}[t]
  \centering
    \includegraphics[width=0.78\textwidth,trim={0cm 0cm 0cm 0cm}, clip]{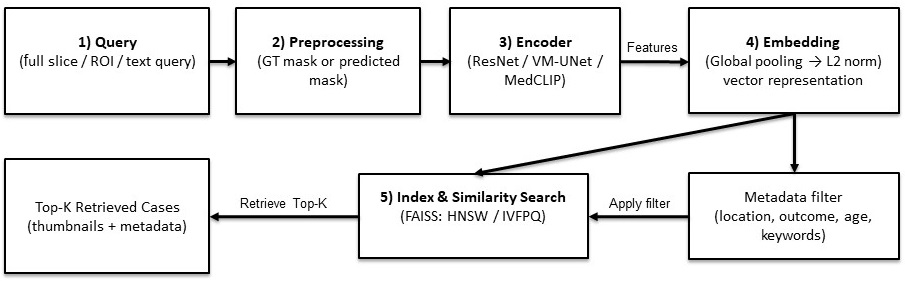}
    \vspace{-3mm}
    \caption{Conceptual prototype for retrieval with AbscessHeNe, illustrating how query images and metadata can be processed and matched to return clinically similar cases.}
    \vspace{-5mm}
    \label{fig:protopype_retrieval}
\end{figure*}

While the current benchmarks focus on segmentation, AbscessHeNe is explicitly structured to support CBIR as illustrated in Figure \ref{fig:protopype_retrieval}. The prototype pipeline begins with a query, which may be a full slice, a region-of-interest (ROI) extracted by ground-truth or predicted masks, or a metadata-augmented request. After optional preprocessing, the image is encoded by a CNN or Transformer backbone, such as ResNet \cite{he2016deep}, VM-UNet \cite{ruan2024vm} encoder, MedCLIP \cite{wang2022medclip}, to generate low-dimensional embeddings through global pooling and normalization. These embeddings are indexed using FAISS \cite{douze2024faiss} for scalable similarity search.
To enhance clinical relevance, structured metadata (e.g., lesion location, multiloculated status, patient age, outcome) can be integrated in two ways: as post-retrieval filters that constrain the search results, or concatenated with image embeddings to form composite queries. Retrieved results are presented as top-K similar cases with thumbnails and metadata links, enabling case-based reasoning.
\section{Discussion}
\label{sec:disc}

\subsection{Clinical and Research Implications}
In this paper, we introduced a novel dataset dedicated to abscess segmentation on head and neck CECT scans. To demonstrate the utility and potential of this dataset, we conducted a series of experiments employing state-of-the-art segmentation models. A total of 4,926 clinically confirmed slices from 47 cases were included, from which 3,935 slices were used for training and 991 slices were reserved for independent testing. Each case included paired CECT images and pixel-wise annotated segmentation masks of abscess lesions, providing comprehensive ground truth references for algorithm development and evaluation.Model performance was assessed using widely adopted evaluation metrics, including DSC, IoU, NSD, and HD95. The best-performing model, VM-UNet, achieved a DSC of 0.39, an IoU of 0.27, an NSD of 0.67, and an HD95 of 21.27 mm, indicating that segmentation performance on the test set remains suboptimal.

These modest results indicate that current models struggle with several failure modes rather than a single bottleneck. As illustrated in Figure \ref{fig:visualize}, the unique challenges posed by head and neck abscesses include variable lesion size, multiloculated morphology, irregular boundaries, and close proximity to critical neurovascular structures. Class imbalance and low contrast at the abscess–soft tissue interface further degrade boundary quality. Therefore, our dataset presents significant challenges for precise delineation, imposing high demands on algorithmic robustness to reliably identify and segment abscess regions in clinical scenarios. In the future, these models can be combined with robust boundary modeling, Hounsfield Unit-aware preprocessing, or explicit anatomical priors. Moreover, we position segmentation as a foundation for content-based indexing and retrieval of head and neck CT. Even with moderate masks, lesion-centered features and structured metadata, such as space involvement, morphology, can support case search, triage, and decision support.

We anticipate that the public release of our abscess dataset will serve as a valuable resource for the medical imaging community, facilitating the development, benchmarking, and validation of more advanced and clinically reliable segmentation algorithms specifically tailored to the complex anatomical landscape of the head and neck region.

\subsection{Limitations and Future Perspectives}
This study has several limitations. First, the dataset used for benchmarking currently comprises 40 training and 7 testing cases, and the evaluation relies on a single train/test split. This limits statistical robustness and generalizability. Moreover, we will expand AbscessHeNe with more cases and richer metadata to better reflect clinical diversity and to enable rigorous content-based indexing and retrieval evaluation, such as Recall@K and mAP, alongside segmentation. These steps aim to improve accuracy, robustness, and interpretability, and to align the system with real-world surgical planning and triage workflows.

Second, segmentation performance remains modest across metrics, especially at lesion boundaries and in multiloculated or low-contrast presentations. To address this, we plan to explore boundary-aware losses, multi-window HU normalization, 2.5D or 3D context, and retrieval-augmented training.

\section{Conclusion}
\label{sec:conc}
In this work, we introduce \textbf{AbscessHeNe}, the first curated and comprehensively annotated dataset of CECT scans dedicated to the semantic segmentation of head and neck abscesses. We establish baseline benchmarks by evaluating a range of state-of-the-art deep learning architectures, offering valuable insights into the current strengths and limitations of automated segmentation approaches in this clinically critical domain. While our results demonstrate the potential of existing models for abscess detection, they also highlight challenges in precisely delineating boundaries, particularly in complex anatomical regions. These findings underscore the need for continued research on more advanced architectures and data-driven learning strategies.
\comment{
AbscessHeNe not only contributes to segmentation but also is structured to support future applications in \textbf{content-based multimedia indexing, semantic annotation, and case-based retrieval}, enabling efficient access to similar clinical cases and enhancing decision support systems. We envision that this dataset will not only drive progress in medical image analysis but also serve as a foundational resource for the development of intelligent, searchable clinical multimedia systems.
}

\section*{Acknowledgment}
  This research is funded by Vietnam National University - Ho Chi Minh City (VNU-HCM) under grant number 36-2024-44-02. 
\balance

\bibliographystyle{IEEEtran}
\bibliography{references}

\end{document}